%% file: icml2025_es-fomo_workshop.tex
\documentclass{article}
\pdfoutput=1

\usepackage{microtype}
\usepackage{graphicx}
\usepackage{subfigure}
\usepackage{booktabs} 

\usepackage{hyperref}

\usepackage[accepted]{icml2025}

\usepackage{amsmath}
\usepackage{amssymb}
\usepackage{mathtools}
\usepackage{amsthm}

\usepackage[capitalize,noabbrev]{cleveref}

\theoremstyle{plain}

\theoremstyle{definition}

\theoremstyle{remark}

\usepackage[textsize=tiny]{todonotes}

\usepackage{comment}
\usepackage{booktabs} 

\usepackage{pdfpages} 

\usepackage{upquote} 
\usepackage{alltt}

\usepackage{csquotes} 

\usepackage{url}

\usepackage{enumitem}
\setlist[enumerate]{left=0pt..1em} 
\setlist[itemize]{align=parleft,left=0pt..1em}   

\icmltitlerunning{GPU Kernel Scientist: An LLM-Driven Framework for Iterative Kernel Optimization}

\begin{document}

\twocolumn[
\icmltitle{GPU Kernel Scientist: An LLM-Driven Framework for Iterative Kernel Optimization}



\icmlsetsymbol{equal}{*}

\begin{icmlauthorlist}
\icmlauthor{Martin Andrews}{sg}
\icmlauthor{Sam Witteveen}{sg}
\end{icmlauthorlist}

\icmlaffiliation{sg}{Singapore}

\icmlcorrespondingauthor{Martin Andrews}{kernelscientist@mdda.net}

\icmlkeywords{Machine Learning, ICML, Kernel, GPU, Optimization, Evolution}

\vskip 0.3in
]



\printAffiliationsAndNotice{}  




\begin{abstract}
%
%
Optimizing GPU kernels for high performance is a complex task, often demanding deep architectural knowledge, extensive profiling, and iterative experimentation. 
This challenge is amplified when targeting newer or less-documented GPU architectures where traditional development aids are scarce. 
This paper introduces an LLM-powered ``GPU Kernel Scientist,'' an automated methodology 
for iteratively refining accelerator kernels. 
%
%

Our methodology employs LLMs in a multi-stage, evolutionary process: 
(a) strategically selecting promising prior code versions as a basis for new iterations; 
(b) generating hypotheses for optimization experiments, 
based on existing code and assimilated knowledge from general GPU literature; and 
(c) autonomously implementing these experiments through code modification 
and subsequent submission to an external evaluation system, using only observed timing data as performance feedback. 
We detail how this approach navigates the challenges of the AMD MI300 target architecture 
and leverages LLMs to compensate for limited domain-specific human expertise. 

In addition to our results, we present the architectural design, operational workflow, and qualitative insights, 
highlighting the potential of LLM-driven agents to democratise and accelerate GPU kernel optimization, 
especially in resource-constrained or rapidly updating hardware environments.

%
\end{abstract}

\section{Introduction}

\begin{figure}
  \centering
  \includegraphics[width=1.0\linewidth]{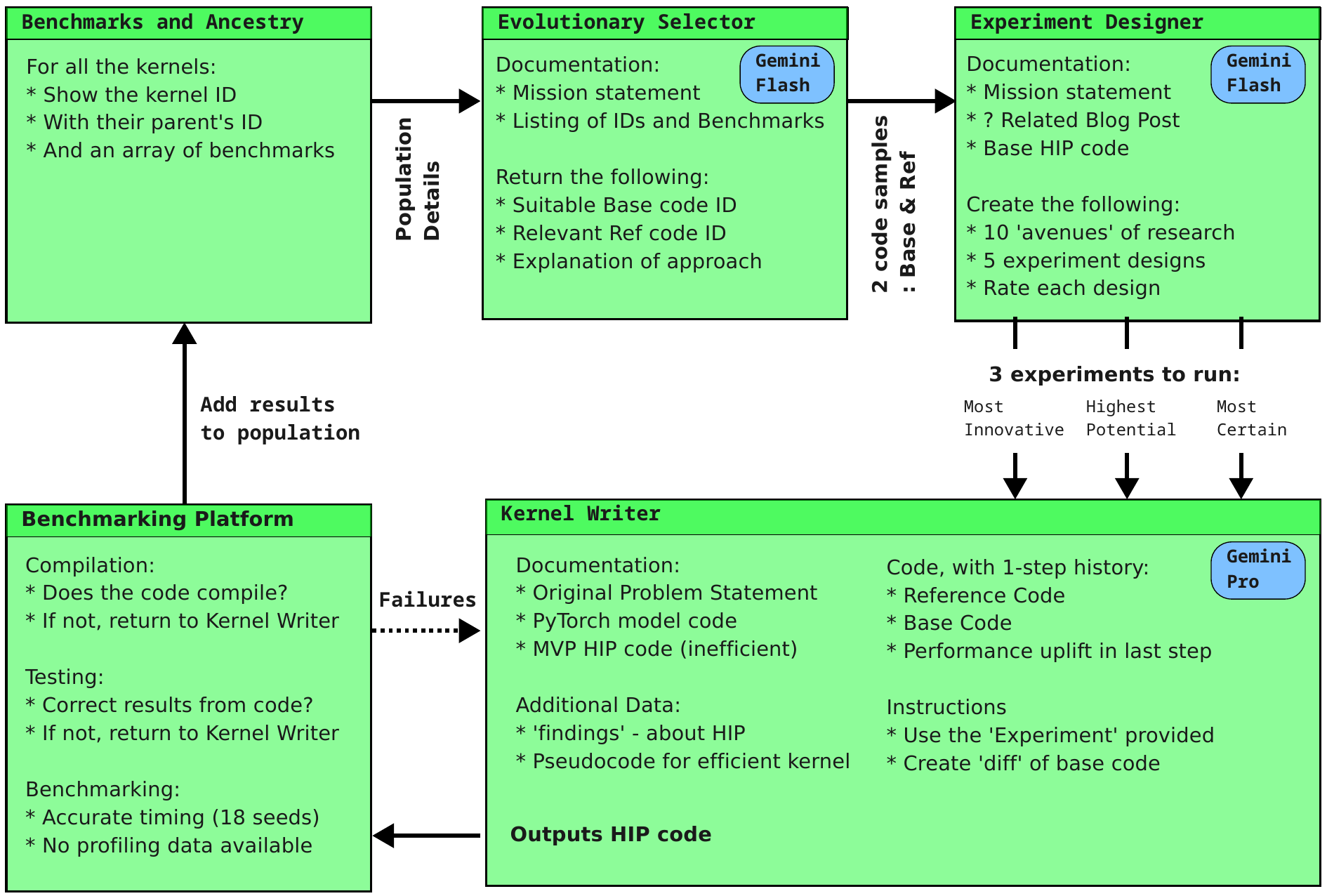}
  \caption{GPU Kernel Scientist Process}
  \label{ScientistProcessDiagram}
\end{figure}



GPU kernel optimization is a significant challenge and traditionally requires specialist expertise.
This challenge becomes magnified when tackling new/niche hardware with limited documentation, 
doubly so with weak profiling tools.

Our proposed solution to this problem is a ``GPU Kernel Scientist'' – an automated, iterative framework
that can optimise kernels for non-CUDA hardware, with access only to end-to-end timing results.

The core idea is to use frontier LLMs
to create a cycle of code selection, experiment ideation, and code generation/modification based only on limited feedback 
from an online testing/benchmark platform.

The key contributions of this brief paper are:
\begin{itemize}[topsep=-3pt]  
\setlength\itemsep{1pt}
\item
Presenting the novel framework itself;
\item
Demonstrating its application to HIP kernel optimization under severe information/tooling constraints; and 
\item
Highlighting how LLMs can bridge knowledge gaps and drive optimization in such scenarios.
\end{itemize}

We will first discuss related work, then detail our methodology, and early experimental findings.  
Examples and code are available in the Appendix and Supplementary materials.



\section{Related Work}
\label{related}


Computer code has long been a target of evolutionary methods \citep{koza1992}, where functions were directly encoded in
tree structures that use analogues of the crossover and mutation analogues from Genetic Algorithms \citep{holland1975} 
to create novel offspring.
In recent times, though, LLMs have provided a new way to create novel individual programs from parent code \citep{alphaevolve, funsearch}.
Here, the LLM acts as both crossover and mutation operator, potentially also tuning its own instructions in the process 
\citep[as in][]{promptbreeder}.

Our approach is to focus on the LLM crossover process, emphasising its active role in the code creation and search process.  
Thus, we have a GPU Kernel Scientist, rather than a GPU Kernel Evolver.

Indeed, this work may also be cast as a synthesis between 
the proposed AI Scientist \citep{2024arXiv240806292L, yamada2025aiscientistv2workshoplevelautomated} 
and the interesting AI CUDA Engineer \citep{lange2025aicudaengineer}.  
While the scope of our `science' is far more limited than the former, 
our focus on writing kernel code is more aligned than the latter.
There are several key differences between this work and the AI CUDA Engineer, where our work  
(i) focusses on one relatively complex kernel (examination of the AI CUDA Engineer outputs shows that it was tackling many simpler tasks);
(ii) optimises for one unusual hardware platform, where the lack of reference material is a significant hurdle; and
(iii) does not have access to profiling information, nor a library of kernels with benchmark data.


The number of challenges available via KernelBench \citep{kernelbench} makes it an excellent benchmark, 
whereas the number of kernels attempted by our system is very limited, 
since our focus is on the kernels required for the The AMD Developer Challenge 2025 \citep{AMDchallenge2025}.
Moreover, the diversity of kernels required for inference of LLMs (for instance) may well be decreasing over time,
while the variety of hardware is increasing : These factors have influenced the design choices made for our GPU Kernel Scientist.




Works such as OpenTuner \citep{opentuner} and KernelTuner \citep{kerneltuner} operate in a way that might be likened to hyper-parameter tuning.
In contrast, the changes to code made by our work can be far more broad in scope - potentially testing out radically different approaches to the kernel optimization problem.
Thus, it seems that these tuning frameworks could also be applied (automatically, via LLM) to the kernels produced by our system,
making the efforts complementary.


%

Frameworks such as OpenAI Triton \citep{triton}, TileLang \citep{tilelang} and ThunderKittens \citep{thunderkittens} 
aim at reducing the detailed work required to implement advanced kernel patterns.  
However, while these frameworks could be a target of future work, the competition environment only has support for Triton, 
and it is not clear the extent to which the AMD low-precision operators would work.  
Thus, we decided in our work to focus on optimising AMD HIP kernels directly.

\section{Methods}
\label{Methods}

Our GPU Kernel Scientist involves the orchestration of 3 LLM 
\footnote{specifically Gemini 2.5 Flash Preview 05-20, and Gemini 2.5 Pro Preview 05-06}
processing stages, 
which iteratively update a growing list of kernels as illustrated in Figure \ref{ScientistProcessDiagram}.

This process was started with a few `seed' kernels, being: 
\begin{itemize}[topsep=-3pt]  
\item 
The (provided) basic PyTorch implementation 
\item
A direct translation of the first kernel into HIP - which was approximately 6 times slower than PyTorch
\item
A translation of the problem into a HIP kernel that made use of the Matrix Cores (\textit{qv} Nvidia's Tensor Cores)
\end{itemize}

While the latter seems like a `gift' to the LLM, the lack of code examples 
using the low-precision operations on this specific hardware
made even human/AI co-creation of a working (verified to give correct results) very challenging.  
This extended deep-dive was then summarised (by the LLM) into a `findings' document, so that the quirks of the hardware could be concisely used by future iterations.

Compared to the extensive CUDA \citep{CUDA} resources available online, 
the AMD documentation is \emph{thin}.
We made use of the 
AMD rocWMMA library \citep{AMDrocWMMA}, 
AMD HIP reference \citep{AMD-MI300-CDNA3-ref},
and 
AMD Matrix Instruction Calculator \citep{AMDmatrixinstructioncalculator}, 
largely by getting the LLM to summarise these sources insofar as they were relevant to the kernel specifications.
%
We also referred the LLM to CUDA resources such as \citet{siboehm-CUDA-MMM}, and \citet{armbruster-fast-matmul} 
which details how Nvidia's Tensor Cores can be used.  The LLM itself `translated' between Nvidia and AMD platforms.


\subsection{LLM Evolutionary Selector}

The first stage in the GPU Kernel Scientist process is to select two individuals from the population of different kernels.
The members of the population are identified by an ID, and the IDs of each of their `parents' is also given, 
as well as the benchmark results for 6 specified MxKxN input configurations.  

The LLM is instructed to choose one of the individuals to be the `Base' for the next experiment, 
and another (the `Reference') to be chosen for its ability to help in analysing experiments.
Thus, instead of having a sophisticated selection mechanism (or other diversity-enhancing ideas, 
such as in \citet{MAPelites}), we rely on the LLM's ability to decide what makes sense 
from the multi-objective optimization situation laid out for it.

See Appendix \ref{appendix-sec1} for examples of the decision process that the LLM reports for several of its choices.

\newpage

\subsection{LLM Experiment Designer}

The second stage in the process is to design useful experiments to be performed subsequently.
For this, we focus on the existing `Base' code for the individual passed in from the prior step,
and augment this code with external knowledge (e.g. blog posts, or summaries of segments of manuals, as described above).

The key instructions here are two-fold:
\begin{itemize}[topsep=-3pt]  
\item 
Firstly, produce 10 `avenues' that might be considered for experimentation - this list is intentionally longer than required, 
but it has been found that this increases the diversity of options available to the second part;
\item 
Secondly, produce 5 experiment plans, which involve an overall experiment description, and then multiple lines of \emph{rubric}.
In addition, the LLM is asked to estimate the range of performance benefit that the experiment might produce, 
as well as the degree to which the experiment is `innovative'.
\end{itemize}


See Appendix \ref{appendix-sec2} for examples of the experiment planning process that the LLM reports.

Following the `planning' of 5 experiments, 3 of them are chosen (without replacement), 
such that the process always choses those experiments that are 
(i) the most innovative; (ii) the highest maximum performance; and (iii) the highest \emph{minimum} performance.
This helps to keep a broad range of alternative paths under consideration, 
and leads to 3 independent instances of the LLM Kernel Writer being launched for the next step in the process.

\subsection{LLM Kernel Writer}


This stage lies at the heart of the GPU Kernel Scientist process : 
Without the ability to produce syntactically correct kernel code, while following the instructions in an experiment's rubric, 
the whole process would clearly fail.
Fortunately, the Gemini 2.5 Pro model used was surprisingly capable of producing valid code, 
even when asked to make significant structural changes.

Extensive context is provided to the LLM:  
\begin{itemize}[topsep=-3pt] 
\item 
Description of the task (as provided by the competition organisers), along with the PyTorch basic implementation for the task;
\item
The `findings' document produced in collaboration with the LLM (see earlier), 
with pseudocode for a high performance kernel;
\item
Two code listings : One for the `Reference' code, the other for the `Base' code. 
The Base code will be the basis of the `diff' through which the output HIP code is produced, 
while the Reference code (suggested by the LLM Evolutionary Selector in the first stage) 
is provided in-context due to its potential to help contrast with or support the code changes.
\item
For each of these code listings, a one-step experiment analysis is given, which consists of the description of the experiment that lead to each
code sample, and the performance benchmarks of its parent and itself.  NB: By construction, all this information will exist.
\end{itemize}

As output, the LLM must produce a new HIP kernel (with calling code, ready for PyTorch integration), 
as well as a short report on which techniques it used to implement the experiment rubric 
(this part was also added to the one-step experiment analysis above, for completeness, since it was occasionally observed 
that the LLM decided against actually following through with the whole experiment rubric).



The Supplementary material (see Appendix \ref{supplementary}) includes an example a HIP kernel produced by the GPU Kernel Scientist, and 
Appendix \ref{appendix-sec3} has a breakdown of its key features.

\subsection{Kernel Testing \& Evaluation}

The kernel testing and evaluation regime was dictated by the 
The AMD Developer Challenge 2025
competition platform.  
Fortunately, command-line tools were available, so the whole process could be automated in a closed loop.

Note, however, that the constraints that the competition's interface 
imposed on our GPU Kernel Scientist also meant
that the system had to be more capable than if those constraints were not present.
The most significant limitation was the lack of any tools for profiling the kernels 
(apart from end-to-end running time for each of the benchmark MxKxN configurations).

In addition, in order to be a `good citizen', 
it was decided that requests for testing/evaluation should only be made sequentially 
(rather than in parallel), 
which limited the overall number of kernels that could be processed.


\section{Findings}

As noted above, there were several constraints around which the GPU Kernel Scientist system had to navigate, 
and these are addressed in the following subsections.

\subsection{Limited Target-Specific Documentation}

The CUDA ecosystem is much better supported than the AMD server-class MI300 GPU targeted for kernel development here.
Thus, the GPU Kernel Scientist system was required to be able to probe the capabilities of the GPU for itself, 
and apply fundamental GPU optimization principles rather than specific hardware pre-trained knowledge.

Also important was the observed ability of the LLM to generalize from related architectures 
(e.g. inferring HIP best practices from CUDA documentation if provided in prompts), 
and then verify its understanding by performing well-chosen experiments.


\subsection{Compensating for Lack of Profiling Tools}

Since the competition submission benchmark timings were the only evaluation tool available, 
the present system had no choice by to use them as the primary performance signal.
Thus, the LLM had to correlate code changes with these black-box timing results 
(implicitly, or by being shown before/after benchmarks).
The authors also observed the LLM performing experiments designed to isolate effects of specific changes to infer their performance impact.

Thus, while having better profiling tools would clearly have been a benefit, 
our GPU Kernel Scientist was shown to be capable of optimising code under harsh constraints.

\subsection{Augmenting Limited Human Expertise}

One clear objective in building the system was for it to perform competitively without the authors 
having to become GPU experts in the process.  
Thus, a conscious decision was made to require all `clever' steps in the process of bootstrapping the system
to originate from the LLM itself.  
This included a lengthy initial hardware probing phase 
(this debugging was done by examining the return values of the competition test interface)
being driven by the LLM, so that it could produce its own `findings' documentation%
\footnote{
Apparently, there were issues concerning the layout of memory blocks on the Matrix Core units, 
and how these could be spread across a warp.  
As may be apparent, the details were left to the LLM, for its own consumption.
}%
.

Once the initial bootstrapping phase was over, all decisions about which code to change, and how, were left to the LLM system.
If new documents were found that might be appropriate (e.g. blog posts), 
they were given to the LLM to digest into a form that was more relevant to the 
current task - and this digested form was made available to the LLM Experiment Designer.

In this way, the LLM became a ``knowledge partner,'' suggesting techniques that the authors 
were not aware of.
In addition, the GPU Kernel Scientist system eliminated any \emph{trial and error} burden for human developers 
by proposing informed experiments that it itself performed iteratively.


\subsection{Iterative Refinement as a Discovery Process}

The loop created for the GPU Kernel Scientist allows the system \emph{as a whole} to learn
about the target architecture through experimentation, without explicit guidance.

It is clear (from reading the output of the LLM Experiment Design process, 
as well as the techniques that the LLM Kernel Writer \emph{chose} to implement) that the 
system can achieve self-consistent directed action through the experimental loop.
Thus, the system has properties that apparently go beyond those of the the individual LLMs that make it up.

\subsection{Competition Results}

Table \ref{Results-table} shows illustrative results from the Competition, where 
the leaderboard was based on the geometric average execution time of the submitted kernel(s) over 
18 specific matrix input sizes.

\begin{table}[!ht]
  \caption{AMD Developer Challenge - summary results}
  \label{Results-table}
  \centering
  \begin{tabular}{lrl}
    \toprule
                      &  Execution  & \\ 
    Implementation    &  time in $\mu$s & Comment \\ 
    \midrule
    PyTorch reference & $\approx 850$ & Uses library fp16 \\
    \midrule
    Human 1\textsuperscript{st} place & $105$ & top-8 had access \\
                        &       & to actual MI300 \\
    \midrule
    Naïve HIP & $\approx 5000$ & Unoptimized \\
    This work & $\approx 450$ & LLM-only \\
    \bottomrule
  \end{tabular}
\end{table}

One unknown about the final leaderboard is the highest position (lowest time) of the human entry 
that \textit{did not have access} to actual hardware.
The authors believe that having access to profiling information, 
and faster iteration cycles for hyper-parameter optimisation,
would enable the LLMs to optimise in a more focussed way.

\section{Discussion \& Future Work}


It is clear that the Gemini 2.5 Pro models can effectively generating syntactically correct and novel HIP code.
Surprisingly, previously common challenges 
(such as ``hallucinations'', producing non-compiling code, or difficulty with highly complex logic)
did not seem to be a major factor.  
This may have been a result of the LLM contexts being carefully controlled, 
with known-working code consistently being present \emph{by construction}.

One surprising facet of this work is the extent to which the system was able to bootstrap itself from very little available documentation.
While the first working HIP kernel was `easy', 
to understand the semantics of the compiler intrinsics for AMD Matrix Cores required
actively probing for compilation/execution errors until the actual behaviour was revealed.
The Gemini 2.5 Pro LLM was found to be capable of prompting for human intervention to enable this kind of debugging process.

In the specific case of the AMD MI300, the kernels developed by humans over the course of the competition 
could provide substantial extra documentation of the capabilities of these GPUs, significantly supplementing
the sample code that was available from public sources.  
This could immediately be used in the context of our work to enable optimisation of other AMD targets.


\subsection{Broader Implications \& Limitations}

The GPU Kernel Scientist system shows potential for democratizing high-performance GPU programming.
This could be used to accelerate kernel development cycles, especially for exploratory work on new hardware,
which is an important factor in an industry that is currently beholden to a single supplier of GPU hardware.

As mentioned earlier, the system's current reliance on external evaluation means that it does not operate in parallel, 
causing it to make slow optimization progress overall.  
In addition, the authors believe that having access to fine-grained feedback 
from profilers would give the GPU Kernel Scientist system a significant boost in capability.


\subsection{Future Work}

The overall structure of the GPU Kernel Scientist lends itself to adaptation to
other target hardware, frameworks and tooling.
Applications could include hardware from vendors with a currently limited installed base, 
or vendors want to build out broader capabilities without having to scale human resources.

To facilitate our system's adaptability to new frameworks 
(where there could potentially be little up-to-date documentation)
we are planning to build automatic tools 
to expand the knowledge base that it can access dynamically.
%
%




In conclusion, our GPU Kernel Scientist demonstrates the profound potential of LLM-driven evolutionary processes 
to navigate and conquer complex optimization challenges, 
even with limited prior knowledge or tooling, 
showcasing a powerful new paradigm for performance engineering.

\section*{Acknowledgements}
\label{Acknowledgements}

Support for this research was provided by the Google AI Developer Programs team, 
including access to the Gemini models and GPUs on Google Cloud Platform.

The authors thank the ES-FoMo III workshop reviewers for their time and valuable feedback.




\bibliography{es-fomo_at_ICML2025}
\bibliographystyle{icml2025}

\newpage
\appendix
\onecolumn

\input{9_appendix}

\end{document}

%% file: 9_appendix.tex
\section{Appendix}

\subsection{LLM Evolutionary Selector}
\label{appendix-sec1}

Here are three typical samples from the output of Evolutionary Selector stage, 
illustrating the way in which the LLM is capable of sophisticated judgements:

{\footnotesize
\begin{verbatim}
basis_code: "00052"
basis_reference: "00046"
rationale: > 
  "Run 00052 is selected as the basis code due to its consistently lowest average 
  benchmark scores across all input configurations, indicating the best overall 
  performance achieved so far. 
  Run 00046 is chosen as the reference because it represents a divergent optimization 
  path from a common ancestor (00037), offering specific strengths 
  (e.g., better performance on the first benchmark than 00052) that can provide valuable 
  comparative insights for the kernel scientist, despite its overall lower performance."
\end{verbatim}
}

{\footnotesize
\begin{verbatim}
basis_code: "00089"
basis_reference: "00087"
rationale: >
  "Run 00089 is selected as the basis code due to its superior overall performance, 
  achieving the lowest average benchmark score and excelling in 
  five out of six configurations. 
  Run 00087, its direct parent, is chosen as the reference because it represents the 
  immediate previous highly optimized iteration, providing crucial context 
  for understanding the precise improvements and minor trade-offs leading to the 
  current best performance."
\end{verbatim}
}

{\footnotesize
\begin{verbatim}
basis_code: "00097"
basis_reference: "00091"
rationale: > 
  "Run 00097 is chosen as the basis for new experiments due to its consistently 
  best overall performance across all benchmark configurations. 
  Run 00091 is selected as the reference because, 
  while an ancestor with a higher total benchmark score, 
  it uniquely performs better on one specific configuration (m=6144, k=512, n=4096), 
  providing valuable insight into optimization trade-offs for the kernel scientist."
\end{verbatim}
}


\newpage
\subsection{LLM Experiment Designer}
\label{appendix-sec2}

Here is a sample from the output of Experiment Designer stage, 
showing how the Avenues (truncated here) provide helpful ideation for the Experiments, 
and also how the LLM believes that it is capable of estimating the likely performance 
gains, and also opining on each experiment's `innovation':

{\footnotesize
\begin{verbatim}
## Task 1: Optimization Avenues

* **LDS Bank Conflict Mitigation for A/B Data:** Analyze and re-pad shared memory ...
* **Optimized LDS Layout for rocWMMA:** Adjust the storage pattern of `lds_a_ping/pong` ...
* **Wider Vectorized Global Loads:** Check if `load_gmem_tile_to_lds_vectorized` ....
* **Increase Thread Block Occupancy:** Explore larger `TBLOCK_X_DIM` values ...
* **Cooperative Store to Global C:** Distribute the final write-back of the C matrix ...
* **Fine-tune Tile Sizes (TB_M, TB_N, TB_K):** Systematically experiment with ...
* **Optimize Scale Application Loop:** Review the unrolled scaling loop for potential ...
* **Asynchronous Scale Loading:** Consider decoupling the loading of scaling factors ...
* **Register Pressure Management:** Profile the kernel's register usage to determine ...
* **Padding Global Memory Inputs:** For smaller matrix dimensions or specific access ...

## Task 2: Experiments

```yaml
experiment:
  - description: >
      "Rectify the LDS data layout for matrix A and B to perfectly match 
      the expectations of `rocwmma::load_matrix_sync` and its fragment types, 
      addressing potential performance bottlenecks from layout mismatches or bank conflicts. 
      The current global-to-LDS load seems to produce a row-major K x M layout for A, 
      while rocWMMA expects column-major M x K for its `MatrixA_MfmaFrag`."
    rubric: >
      "Modify `load_gmem_tile_to_lds_vectorized` to transpose or reorder data during 
      the load from global memory so that `lds_a_target` stores `A` in a column-major 
      M x K format (M being the fastest moving dimension) and `lds_b_target` stores `B` 
      in a row-major K x N format (N being the fastest moving dimension), 
      aligning with the rocWMMA fragment definitions (`col_major` for A, `row_major` for B). 
      Adjust padding if necessary."
    performance: [15, 40]
    innovation: 85
  - description: >
      "Redesign the final C matrix write-back to global memory by distributing 
       the write operations across all active waves in the thread block, rather than just 
       the first wave, to improve global memory write bandwidth utilization and reduce 
       idle time for other waves."
    rubric: >
       "Remove the `if (wave_id_in_block == 0)` guard from the final store loop. 
       Implement a cooperative store pattern (e.g., using `rocwmma::store_matrix_sync` 
       if suitable for `bf16\_t` outputs, or manually distributing the store elements 
       using `threadIdx.x` and `wave_id_in_block`) to ensure all waves contribute to writing 
       the `TB_M x TB_N` tile to global memory. 
       Ensure boundary checks are correctly handled for all contributing threads."
    performance: [5, 15]
    innovation: 60
  - ... etc ...
\end{verbatim}
}

\newpage

\subsection{LLM Kernel Writer}
\label{appendix-sec3}

The following is the output from the LLM when asked to explain the various features of 
a specific kernel produced by the GPU Kernel Scientist process.  
The code for this kernel is included in the Supplementary Material.

The specific prompt used (in conjunction with copy/pasting the HIP code verbatim) was : 
``For the following HIP kernel, please describe the major features that it has (eg: does it do swizzling, use the AMD Matrix Cores, shared memory, etc)''.
The output was then converted to \LaTeX\ format for inclusion here.

\subsection*{\emph{HIP Kernel Major Features (author: Gemini 2.5 Flash)}}

This HIP kernel implements a highly optimized General Matrix Multiply (GEMM) operation ($C = A \times B$) with additional scaling, specifically tailored for AMD GPUs using the \texttt{rocWMMA} library. It leverages various low-level optimizations to maximize performance.

\subsubsection*{AMD Matrix Cores (via rocWMMA)}
\paragraph*{Integration:} The kernel heavily relies on the \texttt{rocwmma} library, which provides C++ interfaces to AMD's Matrix Fused Multiply-Add (MFMA) instructions.
\begin{itemize}
    \item \textbf{Fragment Types:} \texttt{MatrixA\_MfmaFrag}, \texttt{MatrixB\_MfmaFrag}, and \texttt{Accumulator\_MfmaFrag} are defined using \texttt{rocwmma::fragment}, indicating direct utilization of MFMA hardware.
    \item \textbf{MFMA Dimensions:} The kernel is configured for a 32x32x16 MFMA variant (\texttt{MFMA\_M\_TILE\_M = 32u}, \texttt{MFMA\_N\_TILE\_N = 32u}, \texttt{MFMA\_K\_TILE\_K = 16u}), which is typical for \texttt{FP8} inputs on AMD GPUs.
    \item \textbf{Core Operation:} The \texttt{mma\_sync} function from \texttt{rocWMMA} is explicitly called to perform the matrix multiplication and accumulation within a wave.
\end{itemize}

\subsubsection*{Mixed-Precision Arithmetic}
\begin{itemize}
    \item \textbf{Input Data Types:} Input matrices \texttt{A} and \texttt{B} are of type \texttt{fp8\_t} (\texttt{\_\_hip\_fp8\_e4m3\_fnuz}), a low-precision 8-bit floating-point format.
    \item \textbf{Accumulator Data Type:} Intermediate accumulation within the \texttt{Accumulator\_MfmaFrag} is performed in \texttt{float} (single-precision), ensuring higher precision for intermediate sums.
    \item \textbf{Output Data Type:} The final output matrix \texttt{C} is cast to \texttt{bf16\_t} (\texttt{\_\_hip\_bfloat16}), a 16-bit floating-point format. This common pattern (FP8 compute, FP32 accumulate, BF16 output) is prevalent in AI/ML mixed-precision workflows.
\end{itemize}

\subsubsection*{Shared Memory (LDS) and Pipelining}
\paragraph*{LDS Usage:} The kernel extensively utilizes \texttt{\_\_shared\_\_} memory (Local Data Share) for efficient data staging. Specifically, \texttt{lds\_a\_ping}, \texttt{lds\_a\_pong}, \texttt{lds\_b\_ping}, and \texttt{lds\_b\_pong} are declared as shared memory buffers to hold tiles of matrices A and B.
\paragraph*{Double Buffering (Pipelining):} A classic ping-pong double-buffering scheme is implemented for loading A and B tiles into LDS. This design overlaps computation with data transfer, significantly improving occupancy and reducing memory stalls:
\begin{itemize}
    \item While one set of LDS buffers (\texttt{ping} or \texttt{pong}) is used for computation (by \texttt{compute\_lds\_tile\_mma}), the other is concurrently filled with the next tile from global memory (by \texttt{load\_gmem\_tile\_to\_lds\_vectorized}).
    \item Strategic \texttt{synchronize\_workgroup()} calls ensure data coherence and proper synchronization between the load and compute stages of the pipeline.
\end{itemize}

\subsubsection*{LDS Re-purposing for Scale Caching}
\paragraph*{Dynamic Use:} A sophisticated optimization employed is the re-purposing of the \textit{already used} LDS buffers (specifically \texttt{lds\_a\_double\_buffers}) for caching \texttt{a\_scale} and \texttt{b\_scale} values. This occurs \textit{after} the corresponding A/B data has been processed by the MFMA units within that pipeline stage. The technique involves casting the LDS pointers (\texttt{fp8\_t*}) to \texttt{float*} to overlay the scale data.
\paragraph*{Efficiency:} This approach maximizes the utilization of the limited LDS capacity by dynamically allocating different data types to the same memory region at different stages of the kernel's execution. Scales are cooperatively loaded into LDS by all threads in the block.

\subsubsection*{Multi-Wave Thread Block Execution}
\paragraph*{Block Configuration:} The kernel is configured with \texttt{TBLOCK\_X\_DIM = 128u} (or potentially 256u), which is a multiple of AMD's typical wave size (64). This means each thread block launches and manages multiple waves concurrently.
\begin{itemize}
    \item \textbf{Distributed Global Memory Loads:} The \texttt{load\_gmem\_tile\_to\_lds\_vectorized} function distributes the global memory load tasks among \textit{all} threads in the block (\texttt{block\_thread\_id}), ensuring efficient utilization of global memory bandwidth.
    \item \textbf{Redundant Per-Wave MFMA Computation:} Each wave within the block computes the MFMA operations (\texttt{mma\_sync}) \textit{redundantly} on the same LDS data. This is often an efficient pattern for \texttt{rocWMMA} as fragment operations are typically wave-level and results are accumulated within each wave's private registers.
    \item \textbf{Single-Wave Global Memory Write:} Crucially, the final results are written from the accumulator fragments to global memory \textit{only by \texttt{wave\_id\_in\_block == 0}}. This prevents redundant writes and potential write conflicts to the same global memory locations from different waves within the same block, which would otherwise lead to correctness issues or severe performance degradation.
\end{itemize}

\subsubsection*{Scaling and Quantization}
\paragraph*{Dynamic Range:} The kernel takes separate \texttt{global\_a\_scale\_ptr} and \texttt{global\_b\_scale\_ptr} inputs, which are per-matrix-element scaling factors.
\paragraph*{Per-Element Scaling:} It applies a per-element scaling to the accumulated \texttt{FP32} results using the formula \texttt{accum\_value * scale\_a\_val * scale\_b\_val\_inv} (where \texttt{scale\_b\_val\_inv} is the inverse of the B-scale). This scaling is performed \textit{before} the final cast to \texttt{BF16} for storage in global memory. This mechanism is characteristic of dynamic range quantization methods used in machine learning inference to map low-precision inputs to higher-precision outputs.

\subsubsection*{Data Layouts and Access Patterns}
\begin{itemize}
    \item \textbf{Matrix Layouts:} Input matrices \texttt{A} and \texttt{B} are configured for column-major global layout (\texttt{col\_major}), while the output matrix \texttt{C} is configured for row-major global layout (\texttt{row\_major}).
    \item \textbf{Vectorized Loads:} The \texttt{load\_gmem\_tile\_to\_lds\_vectorized} helper function performs vectorized loads from global memory to LDS (e.g., loading 4 \texttt{FP8} elements at once by casting to an \texttt{int*} pointer), optimizing global memory bandwidth utilization.
\end{itemize}

\subsubsection*{Swizzling}
\paragraph*{Implicit Handling:} The kernel does not explicitly implement data swizzling (reordering elements to optimize cache locality or access patterns beyond linear/vectorized loads) within its custom load routines. Instead, this detail is delegated to the \texttt{rocWMMA} library. The \texttt{rocwmma::load\_matrix\_sync} function internally handles the necessary data reordering and packing of matrix fragments to align with the specific memory access patterns and data layouts expected by the underlying MFMA hardware for optimal performance.

\paragraph{}
In summary, this HIP kernel exemplifies a highly optimized, mixed-precision \texttt{GEMM} implementation designed for AMD GPUs, demonstrating advanced techniques for performance and correctness by deeply integrating with hardware features and memory hierarchy.

\subsection{Supplementary Materials}
\label{supplementary}

To access a zip file of Supplementary Materials, please use the following link:

\url{https://drive.google.com/file/d/1uPE6_RLA0agwS2VwF-_ZCwxWoeT1Sr3D/view}